\documentclass{article}
\usepackage{ijcai26}

\usepackage{times}
\usepackage{soul}
\usepackage{url}
\usepackage[hidelinks]{hyperref}
\usepackage[utf8]{inputenc}
\usepackage[small]{caption}
\usepackage{graphicx}
\usepackage{amsmath}
\usepackage{amsthm}
\usepackage{booktabs}
\usepackage{algorithm}
\usepackage{algorithmic}
\usepackage[switch]{lineno}

\usepackage{amssymb}
\usepackage{tabularray}
\usepackage{makecell}
\usepackage{placeins}
\usepackage{placeins}

\title{Generic Interpretation Approach for Transformer Models Incorporating Heterogenous Attention Structures}

\author{
Yongjin Cui$^1$
\and
Xiaohui Fan$^1$\and
Huajun Chen$^{1}$
\affiliations
$^1$Zhejiang University\\
\emails
\{cuiyongjin, fanxh, huajunsir\}@zju.edu.cn
}

\begin{document}

\maketitle

\begin{abstract}
    Transformer has significantly propelled the development of artificial intelligence, and certainly the development of agents as well. We categorize attention structures of Transformer into two types based on the source of the input information: homogenous and heterogenous attention structures. Heterogenous attention structures, with co-attention as a typical example, process information from different sources. Heterogenous attention structure is the foundation for Transformer models to achieve more complex functions and integrate more modal information. Whether for research purposes or policy requirements, the interpretation of Transformer models with heterogenous attention structures is an important task. The fusion of information from different sources brings new challenges. Our work mainly includes two parts: method and experimentation. In terms of method, we propose an interpretation method for Transformer models with heterogenous attention structures. In terms of experimentation, based on our experimental analysis paradigm, we interpret the operating mechanisms of representative models, conduct semantic interpretation and logical interpretation.

\end{abstract}

\section{Introduction}
Transformer is first proposed and applied in the single mode field of natural language processing \cite{1}, and later successfully applied to computer vision \cite{6,7} as well as audio field \cite{13,15}, and now, it is leading multimodal information processing and has achieved significant results \cite{21,30}. 

In these models, we categorize a specific type of attention structure as the homologous attention structure, wherein the information processed by these structures is uniformly sourced from the same information stream, akin to the self-attention structure of Vision Transformer (ViT) \cite{6}. This classification stems from the fact that the Query (Q), Key (K), and Value (V) within this attention structure all hail from the same singular information source. Other attention structures, including co-attention, also called as cross-attention, in multimodal Transformer models or in the decoder of Transformer models, are categorized as the heterogenous attention structure. The reason for making this distinction is that the information source of Q in these structures is different from that of K and V. Notably, the self-attention structure that succeeds the co-attention is also recognized as a unique form of heterogenous attention structure. Despite its Q, K, and V being derived from the same source, this source is the product of the fusion of two distinct information sources through the co-attention process.

Heterogenous attention structures provide an effective way of information fusion and promote the development of multimodal models. Multimodal Transformers not only achieve excellent results in the target task, but also demonstrate exciting emergent abilities. While cutting-edge computer vision models typically undergo training as dedicated systems tailored for specific tasks, predicting a predefined set of labels, \cite{31} achieved unprecedented performance by employing a bi-modal Transformer that creates images that accurately align with given descriptions, even in domains that were not encountered during training. At present, some multimodal large models have gradually been transformed into productivity to serve people's daily life and production, such as GPT-4 and Gemini.

This paper introduces a general interpretation method for heterogenous attention Transformer models, which features clear principle, simple calculation, better flexibility, and better performance. Experiments of this paper are mainly conducted in the field of image and image-text multimodal. Based on our experimental analysis paradigm, we interpret the operating mechanisms of representative models, conduct semantic interpretation and logical interpretation.

\textbf{Our contributions mainly consist of the following two aspects:}
\begin{itemize}
    \item We propose a generic interpretation approach for Transformer models incorporating heterogenous attention structures, which has achieved the best results in experiments.
    \item Our approach can explain the operational mechanisms of typical models, and has enabled semantic interpretation as well as logical interpretation.
\end{itemize}

%%%%%%%%%%%%%%%%%%%%%%%%%%%%%%%%%%%%%%%%%%%%%%%%%%%%%%%%%%%%%%%%%%%%%%%%
\section{Related Work}\label{related}

\begin{figure*}[]
  \centering
  \includegraphics[width=1.7\columnwidth]{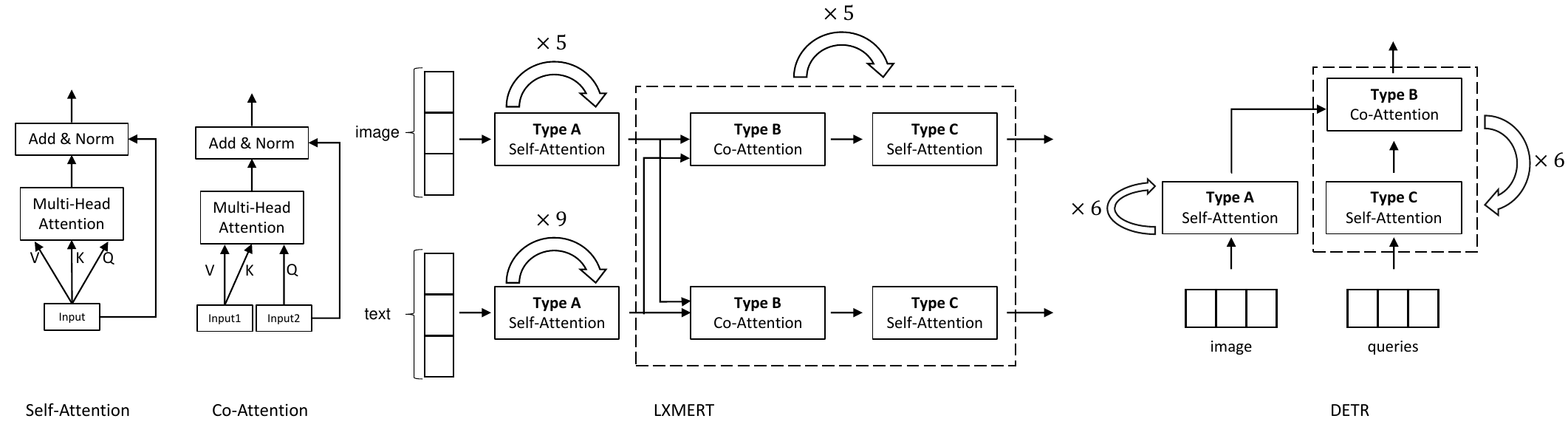}
  \vspace{-3mm}

  \caption{Illustration of Attention Structure Classification.(Homogenous attention structures: $Type A$; Heterogenous attention structures: $Type B$ and $C$)}
  \label{fig1}
    \vspace{-0.7cm}
  \end{figure*}

The interpretation methods of the Transformer model mainly utilize its unique attention structure \cite{1,6}, or combine with the interpretation methods such as GradCAM \cite{35}, LayerCAM \cite{36}, LRP \cite{37}, etc. Vaswani et al. \cite{1} applied the attention of partial layers and partial heads to explain the intrinsic mechanism of Transformer when they first proposed it, and found that different heads perform different tasks. Multi-head mechanism has become a very important issue in model interpretation. Voita et al. \cite{DBLP:conf/acl/VoitaTMST19} proposed partial LRP, suggesting employing the LRP method to assess to what extent different attention heads contribute to the model's prediction instead of considering an average value of attention heads. Michel et al. \cite{57} obtained the same conclusion that different heads perform different tasks and contributed differently, and therefore proposed that pruning the unimportant heads has little impact to the model. Considering information originating from different tokens gets increasingly mixed, making attention weights unreliable as explanations probes, Abnar et al. \cite{58} proposed attention rollout and attention flow to quantify the flow of information through self-attention. And Dosovitskiy et al. \cite{6} applied attention rollout to compute maps of the attention from the output token to the input space when they first proposed Vision Transformer (ViT). Chefer et al. \cite{44} introduced Transformer Attribution (T-Attr) integrating scores throughout the attention graph, by incorporating both LRP-based relevancy and gradient information, in a way that iteratively removes the negative contributions. Chefer et al. \cite{45} introduce Generic Attention-model Explainability (GAE), which combining gradient with multi-head attention maps, and then performing attention rollout. Yuan et al. \cite{yuan2021explaining} explain information flow inside Vision Transformers using Markov Chain (TAM).  Barkan et al. \cite{46} propose Deep Integrated Explanations (DIX), generates interpretation maps by integrating information from the intermediate representations of the model, coupled with their corresponding gradients. Chen et al. \cite{47} propose Beyond Intuition Method (BI) to approximate token contributions inside Transformers, in order to solve  the ambiguity of the expression formulation which can lead to an accumulation of error. Xie et al. \cite{DBLP:conf/ijcai/Xie0CZ23} propose ViT-CX based on patch embeddings, rather than attentions paid to them, and their causal impacts on the model output. Englebert et al. \cite{DBLP:conf/iccvw/EnglebertSNMSCV23} propose Transformer Input Sampling (TIS) a perturbation-based explainability method for Vision Transformers, which computes a saliency map based on perturbations induced by a sampling of the input tokens. Zhao et al. \cite{DBLP:conf/icml/ZhaoWZZC24} propose Grad-ECLIP to interpret Contrastive Language-Image Pre-training (CLIP).Zhao et al. \cite{ODAM} propose gradient-weighted Object Detector Activation Maps (ODAM) to interpret the predictions of object detectors.

\textbf{Baseline Methods.} Among the aforementioned methods, only GAE and ODAM can natively support the interpretation of heterogenous attention Transformers. Among them, ODAM was published in IEEE Transactions on Pattern Analysis and  Machine Intelligence (TPAMI) 2024 and represents the latest research achievement when our work was initiated. Therefore, we selected these two algorithms as our baselines.

\section{Method}\label{sec_method}
% \FloatBarrier

We categorize and illustrate the various attention structures using two exemplary model architectures: DETR \cite{25} and LXMERT \cite{48}, which are also referenced in the work of Chefer et al.\cite{45} for illustrating GAE. DETR is also an example model of ODAM. DETR is a Transformer model for object detection with 6 encoder and 6 decoder layers. LXMERT is a large-scale cross-modality Transformer model learning the vision-and-language connections and consisting of three encoders: 5 object relationship encoder layers, 9 language encoder layers, and 5 cross-modality encoder layers. We classify these attention structures into homogenous and heterogenous categories based on the source of information they process. Homogenous attention structures, labeled as $Type A$ in Figure~\ref{fig1}, process information from the same source, with Q (query), K (key), and V (value) all originating from the same information source. Heterogenous attention structures include two types, labeled as $Type B$ and $C$ in Figure~\ref{fig1}. $Type B$ represents co-attention, also known as cross-attention, where the information sources of K and V differ from that of Q. Although the information source for Q, K, and V in $Type C$ is ostensibly the same, it's derived from the fusion of two information sources through co-attention, making $Type C$ still fall under the category of heterogenous attention structures.

\textbf{Firstly, we solve the problem of multi-head attention.}

\textbf{For both homogenous and heterogenous attention structures}, we correct the attention map $A$ with its gradient to get the average attention map $\bar{A}^{(l)}$ of the $l_{th}$ layer as follows. As for which correction method to use, it will be specifically introduced in the experiment later.

\begin{align}
    \bar{A}^{(l)}=&\mathbb{E}_H\left((\nabla A^{(l)})^{+}\odot A^{(l)}\right) \label{eq1} \\
    or\nonumber\\
    \bar{A}^{(l)}=&\mathbb{E}_H\left(\nabla A^{(l)}\odot A^{(l)}\right) \label{eq2} \\
    or\nonumber\\
    \bar{A}^{(l)}=&\mathbb{E}_H\left(\left|\nabla A^{(l)}\right|\odot A^{(l)}\right) \label{eq3}
\end{align}
Where $\mathbb{E}$ represents the mean operation, $H$ represents number of heads, $l$ represents the $l_{th}$ layer, $\odot$ represents the Hadamard product.

\textbf{Next, we will connect the average attention of different layers as the overall attention.}

\textbf{For homogenous attention structures}, we use attention rollout to get the overall connected attention $\dot{A}$:
\begin{equation}
    \tilde{A}^{(l)}=I+\bar{A}^{(l)}
\end{equation}
\begin{equation}
    \dot{A}^{(l)}=\tilde{A}^{(l)}\cdot\tilde{A}^{(l-1)}\cdot...\cdot\tilde{A}^{(2)}\cdot\tilde{A}^{(1)}
\end{equation}
Where $\tilde{A}^{(l)}$ represents equivalent attention map of the $l_{th}$ layer, $I$ represents the identity matrix to equivalent residual connection, $\dot{A}$ represents the connected attention map, $\cdot$ represents matrix multiplication.

\textbf{For heterogenous attention structures}, our guiding principle is to consistently and linearly distinguish between different information sources. For a certain layer and a certain heterogenous attention structure, the calculation method is as follows.
\begin{equation}
    A\_out^{(l)}{}_{source1}=A\_in\_q^{(l)}{}_{source1}+\tilde{A}^{(l)}\cdot A\_in\_v^{(l)}{}_{source1}
\end{equation}
\begin{equation}
    A\_out^{(l)}{}_{source2}=A\_in\_q^{(l)}{}_{source2}+\tilde{A}^{(l)}\cdot A\_in\_v^{(l)}{}_{source2}
\end{equation}
Where $A$ represents attention scores, the suffix $out$ and $in$ represent the output and input of heterogenous attention structures respectively, the suffix $q$ and $v$ represent the Query and Value, the subscript $source1$ and $source1$ represent the two information sources.

\textbf{Finally,} after the above calculation, we can obtain the final attention score matrix $A\_out_{source1}$ and $A\_out_{source2}$. We only take the row of $CLS$ as the interpretation.
% \FloatBarrier

\section{Comparison with GAE}\label{GAE}
% \FloatBarrier

We conducted a detailed investigation of 10 transformer interpretation algorithms in Section \ref{related} from 2020 to 2024, among which only only GAE and ODAM can natively support the interpretation of heterogenous attention Transformers. ODAM is a method based on intermediate variables. Both our method and GAE are based on attention maps. Here, we compare our method with GAE.

The idea of GAE is similar to T-Attr proposed by the author before. Both of them start from the perspective of relevance. T-Attr continuously assigns relevance from the output to the input side, while GAE accumulates relevance from the input to the output side. Moreover, GAE considers the interaction of different signal source information, which is specifically manifested in Equation (8), (9) and (10) \textbf{in GAE \cite{45} paper}. However, Equation (10) \textbf{in GAE \cite{45} paper} is only defined without explaining the rationality of this definition.

Our method follows this line of thinking. In the transformer model, both the encoder and decoder are feature extraction layers, which ultimately pass the features to relevant decision-making layers, such as classifiers. The role of the decision-making layer can be reflected in the feature extraction layer through the gradient. So, we only need to approximate the feature extraction process as accurately as possible to get the overall attention to the input as the interpretion result. Moreover, we believe that the interaction of different source information is already reflected in the attention map. Therefore, we do not focus on the interaction between different information sources in the heterogenous attention structure, but only on the amount of information from different information sources in the final features. In other words, during our interpretation process, we always maintain the independence between the information from the two sources. This is completely different from GAE.

Although our method is different from GAE in terms of thinking, when we adopt the \textbf{positive} gradient correction, there is a certain similarity between our method and GAE in terms of implementation. When we adopt the \textbf{positive} correction and remove Equation (8) and (9) and replace Equation (10) in GAE with $R^{qk}=R^{qk}+\bar{A} \cdot R^{kk}$ in all layers, our method and GAE are equivalent in implementation.

\textbf{Our method performs better.} We point out the differences in thinking and similarities in implementation here, and subsequent experiments prove that our method performs better, as shown in Figure~\ref{fig2}~and~\ref{fig3}, Table \ref{lingdiansanbeitable}.

\textbf{Our method is clearer in principle.} There is no reasonable basis to explain the rationality of the interaction between different information sources in GAE. Later experiments also prove that this part of information interaction actually brings more noise, as shown in Figure~\ref{fig2}~and~\ref{fig3}.

\textbf{The calculation of our method is simpler.} Our method avoids the computation of GAE Equation (8) and (9), and one time matrix multiplication calculation in GAE Equation (10), in the interpretation process of all heterogenous attention structure layers.

\textbf{Our method is more flexible.} Our method can achieve complete gradient correction (Equation \ref{eq2}) and absolute gradient correction (Equation \ref{eq3}), which is shown in Figure \ref{fig2}, \ref{fig3} and \ref{fig6}, and will be detailed in subsequent experiments.

We divide the main ideas of model interpretation methods into two categories. One is the reductionist method, which emphasizes the interpretation of each step and only focuses on positive contributions, such as GAE and T-Attr. Another is the holistic approach, which emphasizes the synergy of all aspects, rather than the interpretation of individual aspects without considering the whole. When we adopt complete gradient correction in our method, we simultaneously retain both positive and negative contributions in each step, and rely on the synergistic effects of all steps to obtain results. This is an embodiment of the holistic approach.

% \FloatBarrier

\section{Experiment and results}\label{experiment}
% \FloatBarrier

All our experiments are conducted on a NVIDIA A100-SXM4-80GB GPU.

\textbf{Baseline methods.} GAE and ODAM. 

\textbf{Sample cases.} For the sake of fairness in comparison, we use the cases available in the GAE and ODAM paper to avoid any suspicion of selecting cases for experimentation.

\textbf{Sample models.} The models, DETR and LXMERT, used in the following experiments are also the same two used in the GAE paper. DETR is also used in ODAM paper.

\textbf{Ours (abs)} represents our method with absolute gradient correction. \textbf{Ours (pos)} represents our method with positive gradient correction. \textbf{Ours (noised)} represents our method with positive gradient correction and a noise link (Equation \ref{eq8}, \ref{eq9} and \ref{eq10}).

In this part, we evaluate the method effect and introduce an experimental paradigm based on loss design and  gradient correction for attention map, which achieves text semantic interpretation, model working mechanism inspection, and logical inspection.
% \FloatBarrier

\subsection{DETR Interpretation}

\begin{figure}[]
    \begin{center}
    \centerline{\includegraphics[width=0.8\columnwidth]{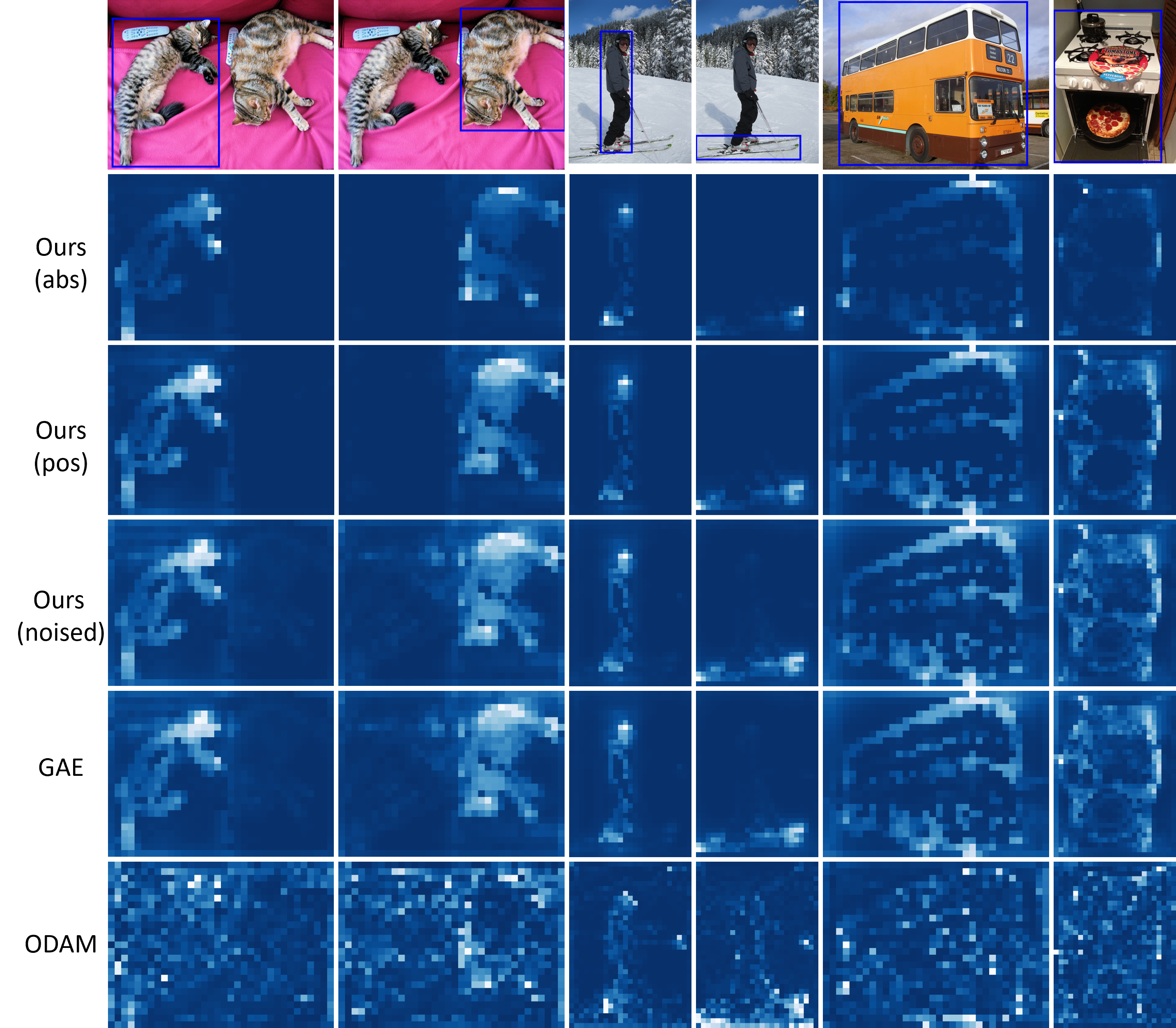}}
    \vspace{-3mm}

    \caption{Sample segmentation masks for DETR. Each row represents a method. Detections (from left to right): cat, cat, person, skis, bus, oven.}
    \label{fig2}
    \end{center}
    \vspace{-0.8cm}

  \end{figure}
  
  \begin{figure}[]
    \begin{center}
    \centerline{\includegraphics[width=0.8\columnwidth]{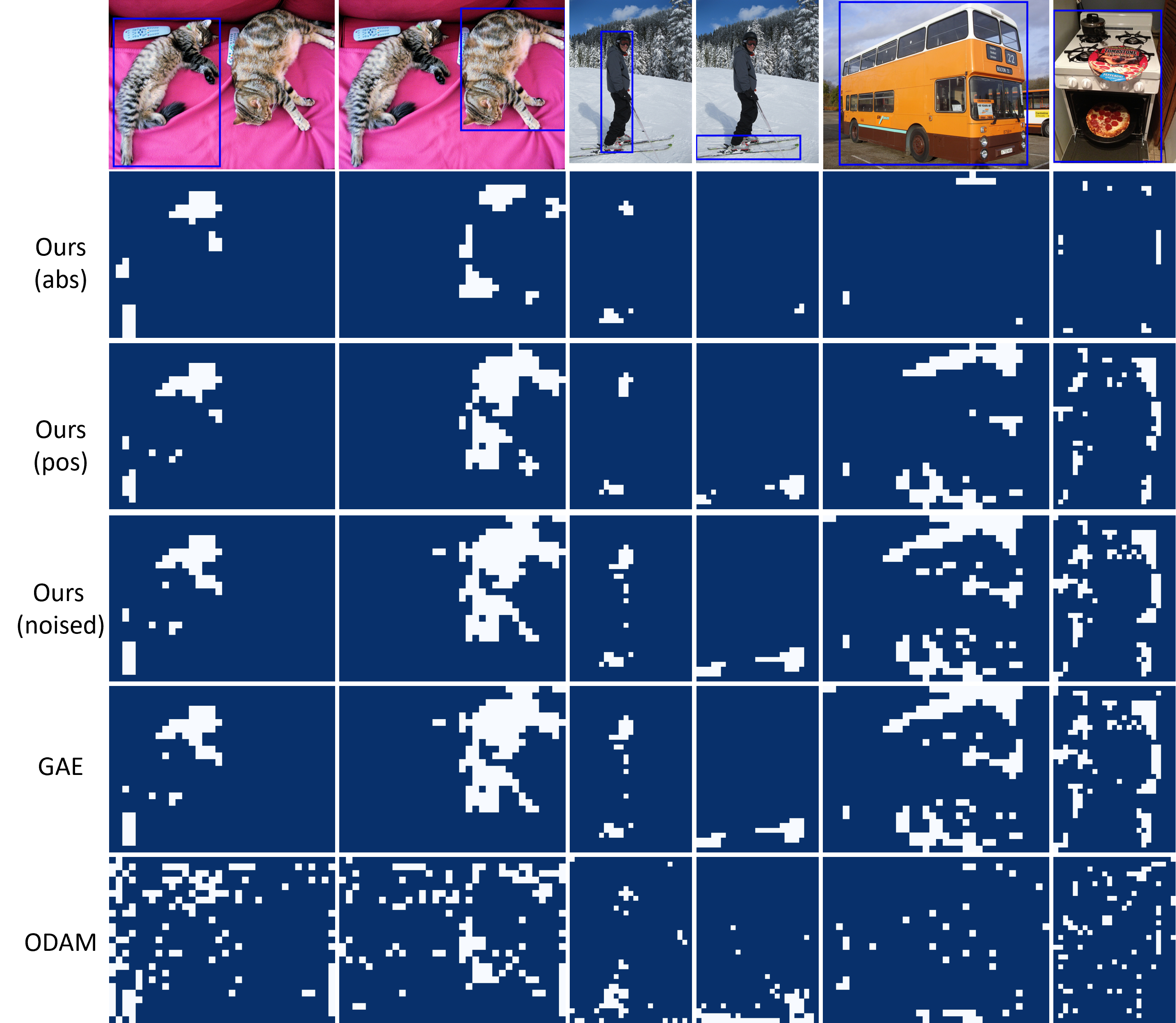}}
    \vspace{-3mm}

    \caption{Sample segmentation mask of DETR after binarization by Otsu method.(The original interpretation result has been destroyed.)}
    \label{yibei}
    \end{center}
    \vspace{-0.8cm}

    \end{figure}
  
  \begin{figure}[]
    \begin{center}
    \centerline{\includegraphics[width=0.8\columnwidth]{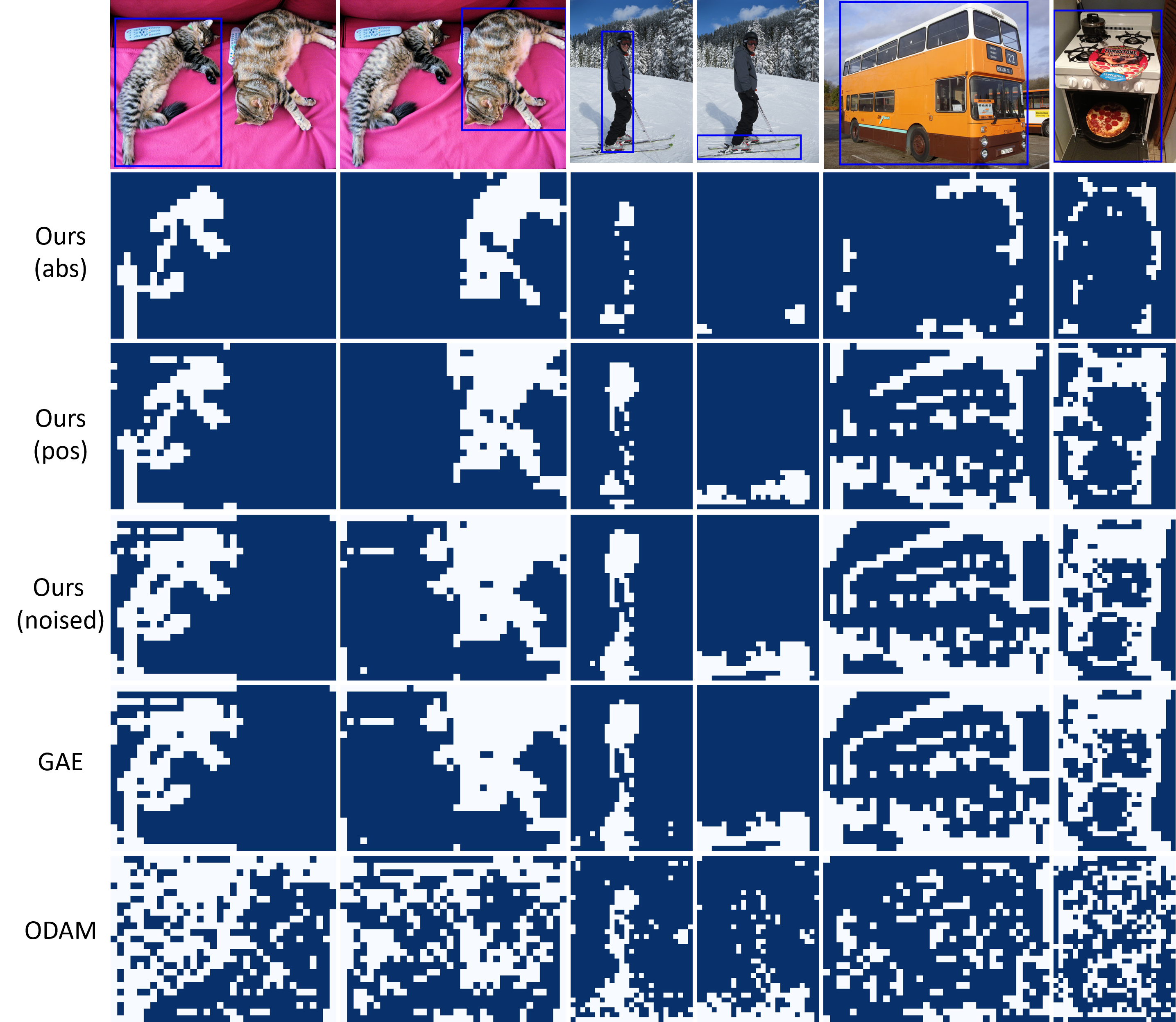}}
    \vspace{-3mm}

    \caption{Sample segmentation mask of DETR after binarization by 0.3 times the Otsu method threshold.(The original interpretation result has been retained.)}
    \label{lingdiansanbei}
    \end{center}
    \vspace{-0.8cm}

    \end{figure}

The DETR model can fulfill both object detection and classification tasks. For the experimentation on this model, we followed the experimental design of GAE. Firstly, we conducted a visualization presentation, visualizing the interpretation results of the model for image tokens. It should be noted that the example images we used in the visualization experiment are all consistent with GAE, and were not selected deliberately. Then, we conduct the same batch experiment as in the GAE paper, which is conducted on the 5,000 samples of the MSCOCO~\cite{49} validation set, with the goal of using the model's interpretation results as a mask for evaluating the results. We first filter the queries to retain only those with a classification probability exceeding 50\%, then employ the Otsu's thresholding method~\cite{50} to separate the foreground and background in the interpretation results. Finally, we upsample the results to obtain the final mask of the original image size and evaluate the mask. During the evaluation, we decrease the minimal IoU (Intersection over Union) used for MSCOCO evaluation from 0.5 to 0.2, as the generated mask, especially after Otsu processing, is often discontinuous.

\begin{table}[]
  \caption{DETR-based weakly supervised segmentation results on the MSCOCO validation set when the threshold is the Otsu method threshold. AP=average precision, AR=average recall. The subscripts denote different benchmark subsets.}
  \vspace{-0.3cm}

  \label{table1}
  \begin{tabular}{lrrrrr}\toprule
    & \textit{GAE}  & \textit{ODAM} & \makecell{\textit{Ours} \\ \textit{(abs)}}& \makecell{\textit{Ours} \\ \textit{(pos)}} & \makecell{\textit{Ours} \\ \textit{(noised)}} \\ \midrule
    $AP$       & 13.1 & 1.9  & 8.0                   & 11.6                 & \textbf{13.2}               \\
    $AP_{medium}$ & 14.4 & 3.3  & 12.2                 & 13.9                 & \textbf{14.4}               \\
    $AP_{large}$  & 24.6 & 2.7  & 12.3                 & 20.8                 & \textbf{24.8}               \\
    $AR$       & 19.3 & 4.8  & 14.7                 & 18.0                   & \textbf{19.4}               \\
    $AR_{medium}$ & 23.9 & 8.0  & 23.6                 & 23.8                 & \textbf{24.0}               \\
    $AR_{large}$  & 33.2 & 6.4  & 19.5                 & 29.2                 & \textbf{33.3}               \\ 
    \bottomrule
  \end{tabular}
  \vspace{-0.5cm}

\end{table}

\begin{table}[]
  \caption{DETR-based weakly supervised segmentation results on the MSCOCO validation set when the threshold is 0.3 times the Otsu method threshold.}
  \vspace{-0.3cm}

  \label{lingdiansanbeitable}
  \begin{tabular}{lrrrrr}\toprule
    & \textit{GAE}  & \textit{ODAM} & \makecell{\textit{Ours} \\ \textit{(abs)}}& \makecell{\textit{Ours} \\ \textit{(pos)}} & \makecell{\textit{Ours} \\ \textit{(noised)}} \\ \midrule
    $AP$       & 7.1 & 2.2  & \textbf{9.1}                  & \textbf{8.9}                 &7.3               \\
    $AP_{medium}$ & 4.6 &1.3  & \textbf{8.3}                 & \textbf{6.2}                 & 4.8               \\
    $AP_{large}$  & 14.1 & 4.3  & \textbf{18.8}                 & \textbf{18.0}                 & 14.5               \\
    $AR$       & 11.3 &4.6  & \textbf{14.5}                 & \textbf{13.6}                  & 11.7               \\
    $AR_{medium}$ & 7.8 & 2.2  & \textbf{13.3}                 & \textbf{10.3}                 & 8.2               \\
    $AR_{large}$  & 26.7 & 11.6  & \textbf{32.1}                 & \textbf{31.6}                 & 27.3               \\ 
    \bottomrule
  \end{tabular}
  \vspace{-0.5cm}

\end{table}

As shown in Figure~\ref{fig2}, the results obtained by ODAM are the worst. Ours(abs) and ours(pos) provide the most accurate visualization results. The results obtained by GAE contain more noise. For instance, when explaining ``cat" using GAE, a faint highlight of another cat's area can be observed. When explaining ``oven", ours(abs) and ours(pos) cleanly remove the pizza area, whereas the GAE method still highlights the pizza area. But in Table~\ref{table1}, GAE gets better scores than our methods, which contradicts the experimental results in Figure~\ref{fig2}.

\textbf{Noise Link}

We believe that the accuracy of evaluation indicators in quantitative experiments is limited. We add a \textbf{noise link} to our method. We perform the following processing on the output of the encoder during the DETR interpretation.

\begin{equation}\label{eq8}
    A\_{add}=A\_out-I
\end{equation}
\begin{equation}\label{eq9}
    S=\sum A_{add}
\end{equation}
\begin{equation}\label{eq10}
    A_{noised}=\frac{A_{add}}{S}+I
\end{equation}

The role of this noise link is the same as the Equation (8) and (9) in GAE paper actually.

In Figure~\ref{fig2}, ours(noised) gets a result looking exactly like GAE's. In Table~\ref{table1}, ours(noised) get the best performance. In other words, noise makes quantitative evaluation results better. The quantitative evaluation results do not have much reference value. It also shows that the relevant operations in Equation (8), (9) and (10) in the GAE paper are not rigorously reasonable, but instead introduce noise. 

\textbf{Why does this situation occur?} It's because the Otsu's method used in the evaluation process is more advantageous in scenarios with more noise, as shown in Figure \ref{yibei}.

We set the binarization threshold to 0.3 times the Otsu's threshold for the experiment. At 0.3 times the threshold, the binarization result is most similar to that in Figure \ref{fig2}, and can reflect the actual interpretation result to a greater extent. The experimental results are shown in Figure \ref{lingdiansanbei} and Table \ref{lingdiansanbeitable}. Ours(abs) gets the best results, followed by ours(pos).

This experiment illustrates that there is a certain bias and error in the evaluation metrics used in batch experiments, which is also reflected \textbf{in the GAE paper}. Figure 4 \textbf{in the GAE paper} demonstrates that GAE performs better than T-Attr. However, in the quantitative experiments, Figure 3(b) shows that T-Attr actually performs better. This is also the same case for our experiment in Figure \ref{fig3} and \ref{fig7}.

Absolute gradient correction only reflects the attention of the model, without distinguishing between categories. Figure~\ref{fig2} ours(abs) shows that the feature extraction process of DETR only filters target object, rather than extracting all object features to further filter target object. 

% \FloatBarrier

\subsection{LXMERT Interpretation} 
% \FloatBarrier

\begin{figure}[]
    \begin{center}
    \centerline{\includegraphics[width=1\columnwidth]{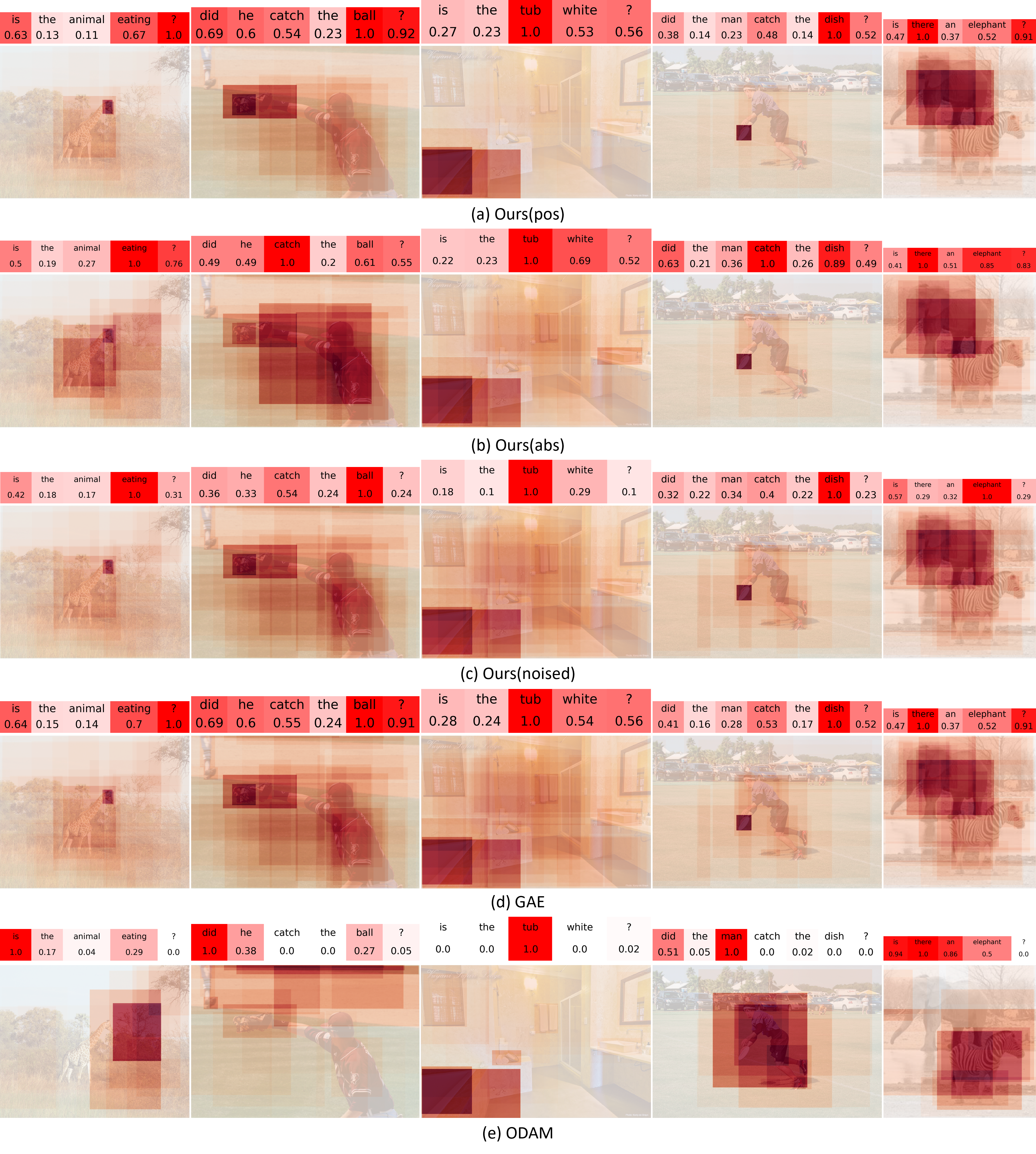}}
    \vspace{-3mm}

    \caption{Interpretation of LXMERT in the VQA task. Answers (from left to right): no, yes, yes, no, yes. (Although the interpretation results are very similar, a careful comparison reveals that ours(pos) has less background noise in the interpretation of every image.)}
    \label{fig3}
    \end{center}
    \vspace{-0.8cm}
    \end{figure}

The results are shown in Figure~\ref{fig3}. ODAM gets the worst results. Ours(pos) and GAE achieved similar results, but a detailed comparison shows that in the interpretation of image content, our method has less noise in the interpretation results. For example, in the interpretation of the third image, our method assigns less contributions to the room background than GAE. Upon closer inspection, other cases exhibit the same pattern. Other methods get more noise.

\begin{figure*}[]
    \begin{center}
    \centerline{\includegraphics[width=2\columnwidth]{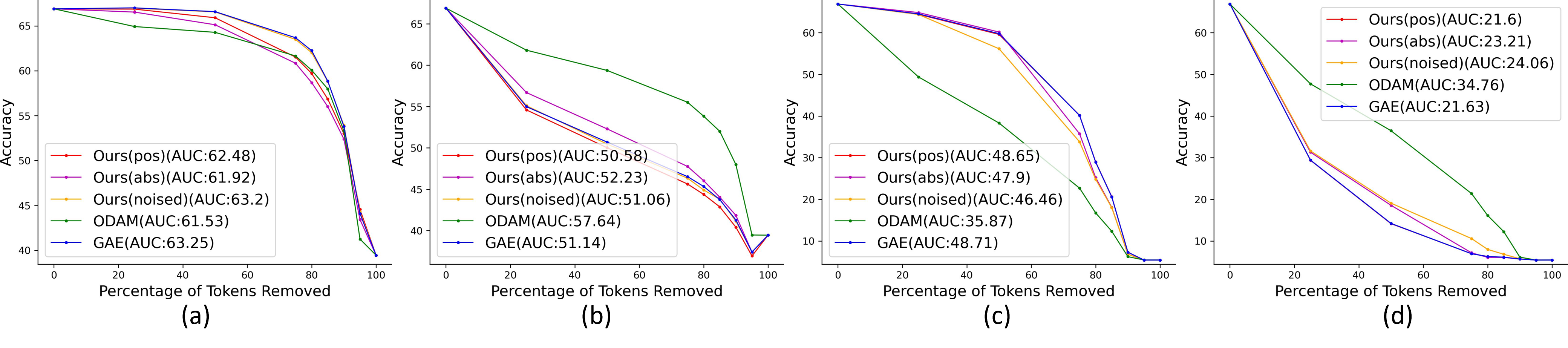}}
    \vspace{-3mm}

    \caption{LXMERT perturbation test results. For negative perturbation, larger AUC is better; for positive perturbation, smaller AUC is better. (a) negative perturbation on image tokens, (b) positive perturbation on image tokens, (c) negative perturbation on text tokens, and (d) positive perturbation on text tokens. (Due to the limitations of quantitative experiments, although our methods is more accurate, it is not fully reflected in quantitative experiments, especially in negative perturbation experiment on image tokens. A similar contradictory situation also exists in GAE paper.)}
    \label{fig7}
    \end{center}
    \vspace{-0.8cm}

    \end{figure*}

Although the previous experiments have demonstrated the limitations of quantitative experiments, which may lead to inaccurate results, for the sake of experimental integrity, we still provide quantitative experimental results (Figure~\ref{fig7}) for reference only. We adopt the positive and negative perturbation experiments in the visual question answering (VQA) \cite{51} task. The process for conducting both positive and negative perturbation tests is outlined below: Initially, a pre-trained network is employed to extract attention score matrix for a randomly selected subset of 10,000 samples from the validation set of the VQA dataset. Next, we systematically remove tokens of a specified modality and assess the network's mean top-1 accuracy. In the case of positive perturbation, tokens are removed starting from the highest score to the lowest. Conversely, in negative perturbation, tokens are removed from the lowest score to the highest. During positive perturbation, a sharp decline in performance is anticipated, suggesting that the removed tokens are crucial for the classification score. On the other hand, negative perturbation aims to maintain the model's accuracy by removing tokens unrelated to the classification. In both scenarios, we calculate the area under the curve (AUC) to quantify the reduction in the model's accuracy.

There is one problem that we cannot directly evaluate the interpretation results of text parts as we can for image interpretation results, because the final interpretation results of text integrate both semantic and grammatical content. However, through our method's better performance in the interpretation of DETR and LXMERT images, it can be reasonably inferred that our interpretation results on text are also better. Because there is no essential difference in the processing flow of data for these two modalities during the method processing.

\textbf{Next, we will conduct the interpretation of semantic and basic image features.}

In the model, the heterogenous attention part is actually a process of mutual query and alignment between two signal sources, while the homogenous attention part is the feature extraction part. The text features output by the homologous attention part will highlight the semantic focus to be queried in the image. The heterogenous attention part integrates the grammatical information of the text while querying. The figure features output by the homologous attention part will highlight basic figure regions.

Therefore, we employ the absolute gradient correction scheme to get the interpretation of the homologous attention part—capturing both the semantic focus and all objects identified within the image.

\begin{figure*}[]
    \begin{center}
    \centerline{\includegraphics[width=1.8\columnwidth]{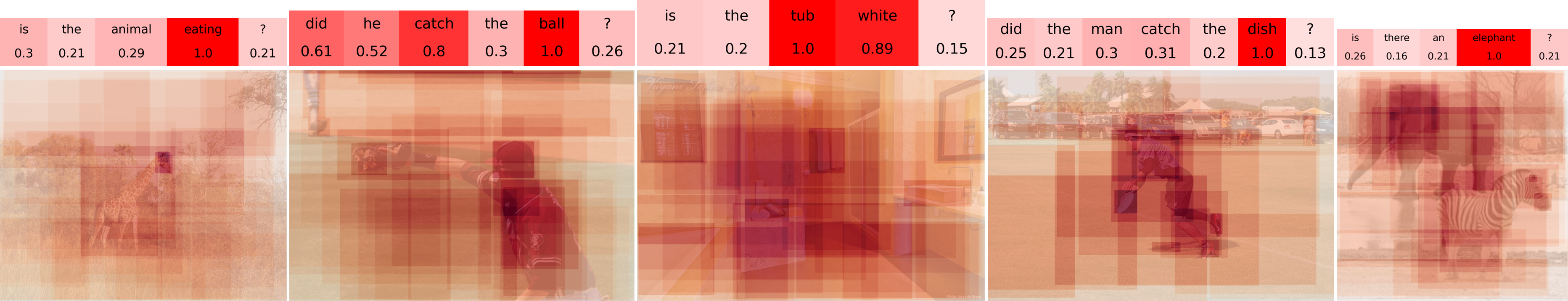}}
    \vspace{-3mm}

    \caption{Text semantic interpretation experiment.(The semantic content of the text is highlighted, and the featur regions of the images are also emphasized.)}
    \label{fig4}
    \end{center}
    \vspace{-0.8cm}

    \end{figure*}

Since the cls token in LXMERT is in the text encoding, it is possible to directly intercept the homogenous attention interpretation output's cls token feature as the semantic interpretation part of the text. Since the image part is not associated with the CLS token during the homologous attention process, We cannot interpret the image's homologous attention component in the same way as we interpret textual homologous attention. We use the total attention obtained by each patch in the image's partially homologous attention output as the features extracted from the image's homologous attention component. The experimental results are shown in Figure \ref{fig4}. The interpretation of text highlights the semantic focus, while the interpretation of images emphasizes the various target subjects within the picture. This also reflects the working mechanisms of LXMERT: LXMERT first separately extracts features from the information of two modalities, and then continuously aligns them in subsequent processes.

\textbf{Logical Inspection.}

\begin{figure*}[]
    \begin{center}
    \centerline{\includegraphics[width=1.8\columnwidth]{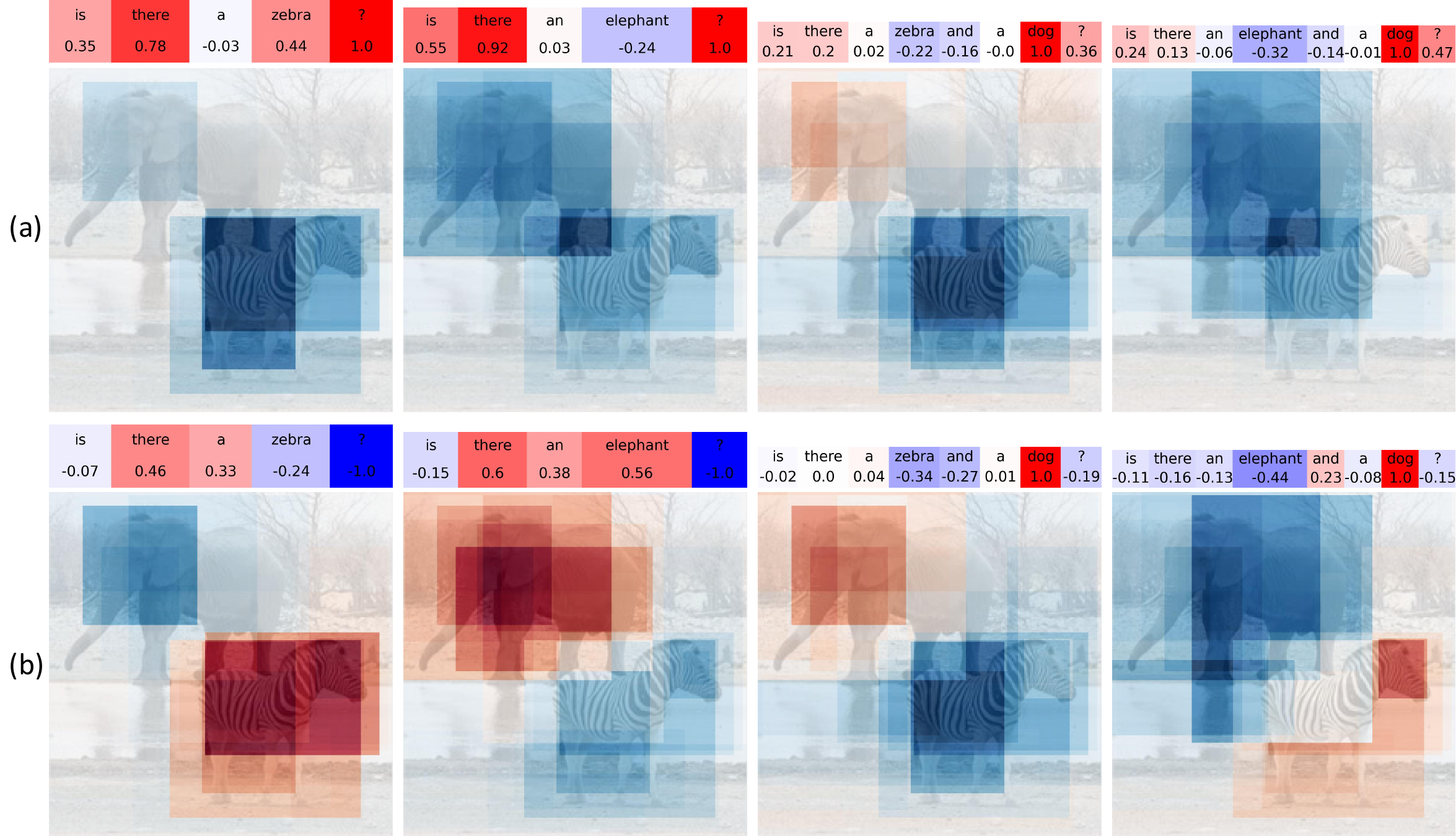}}
    \vspace{-3mm}
    \caption{Verify the working mechanism of LXMERT and the logical verification of the image content under the full gradient correction scheme. Answers (from left to right): yes, yes, no, no. $loss$ of (a) (from left to right): $logit_{yes}$, $logit_{yes}$, $logit_{no}$, $logit_{no}$. $loss$ of (b) (from left to right): $logit_{yes}-logit_{no}$, $logit_{yes}-logit_{no}$, $logit_{no}-logit_{yes}$, $logit_{no}-logit_{yes}$.}
    \label{fig6}
    \end{center}
    \vspace{-0.8cm}

    \end{figure*}

Our method is more flexible than GAE, allowing complete gradient correction. Under complete gradient correction, our method preserves more information, enabling better validation of the logic behind the model's judgments. The interpretation results are shown in Figure \ref{fig6}. In Figure \ref{fig6} (a), we set the loss as the logit of the output to be interpreted, but the interpretation results are not satisfactory. In Figure \ref{fig6} (b), we set the loss as the difference between the logits of two answers that conform to grammatical and semantic correctness, and the interpretation results fully align with the logic. 

When we set the loss as the logit of one output, the model has multiple ways to reduce this output. Ultimately, the direction of the gradient is jointly determined by the target output and other outputs, making the interpretation results difficult to comprehend. However, when we set the logit in the manner depicted in Figure \ref{fig6} (b), the model is more inclined to shift its attention from the feature regions of $output1$ to those of $output2$, that is, decreasing {logit1} while increasing {logit2}. This facilitates our assessment of whether the model has grasped the correct logic. In Figure \ref{fig6} (b), the red regions represent the feature regions of $output1$, and the blue regions represent those of {output2}. We can verify the correctness of the model's logic. For instance, when asking "is there a zebra?", the red zebra regions tend to yield a "yes" answer, while the blue elephant regions tend to yield a "no" answer. When asking "Is there a zebra and a dog?", the blue zebra regions, which align with the text, tend to give a "yes" answer, but the red elephant regions are judged by the model not to be a dog, thus leaning towards a "no" answer. The logical analysis for other cases follows the same pattern.

For the text Interpretation in Figure \ref{fig6}, since it is a blend of grammar and semantics, it is challenging for us to grasp intuitively. Just as we cannot directly comprehend the results of the text interpretation in Figure \ref{fig3}.

In addition, the setting of the loss function is primarily aimed at guiding the gradient. We set the loss as $loss = logit_1 - logit_2$, or $loss = logit_1 / logit_2$  , or enhance it to $loss = (logit_1 - logit_2) / logit_2$. Any kind of loss function is acceptable as long as it can effectively guide the gradient.

\section{Conclusion}\label{conclusion}
% \balance
% \FloatBarrier

As the performance of models improves, the connection between artificial intelligence and people's daily life is becoming closer and closer. We should place equal importance on model interpretation and model performance to ensure the safety and credibility of AI. Heterogenous attention structure is the foundation for Transformer models to achieve more complex functions and integrate more modal information. Whether for research purposes or policy requirements, the interpretation of Transformer models with heterogenous attention structures is an important task. However, there is still limited research on the interpretation of such models, and people's attention is more focused on model performance. The fusion of information from different sources also brings new challenges to model interpretation.This paper proposes a novel interpretation method for Transformer models with heterogenous attention structures, which features clearer principle, simpler calculation, better flexibility. Additionally, based on our method and experimental paradigm, we achieved semantic interpretation and logical interpretation of the model. This study conducts analysis based on two typical models, and continuous research is still needed for more models and more complex attention structures. The interpretation of grammar and semantics in text will also be our key focus going forward.

\FloatBarrier

\bibliographystyle{named}
\bibliography{duomotai}

\end{document}